\documentclass[12pt,a4paper,twoside]{article}
\usepackage[utf8]{inputenc}
\usepackage[T1]{fontenc}
\usepackage{hyphsubst}
\usepackage[margin=2.5cm]{geometry}
\usepackage{lmodern}
\usepackage[square, comma, numbers, sort&compress, round]{natbib}
\usepackage[english]{babel}
\usepackage{mathtools, nccmath}
\usepackage{bef_aras_nothm,todonotes}
\usepackage{amsmath,amsthm,amssymb}
\usepackage{fixmath}
\usepackage{upgreek}
\usepackage{amsfonts}
\usepackage{microtype}
\usepackage{float}
\usepackage{mathrsfs}
\usepackage{hyperref}
\usepackage{paralist}
\hypersetup{
     colorlinks   = true,
     citecolor    = blue
}
\usepackage[nameinlink]{cleveref} 
\crefname{figure}{\figurename}{\figurename}  
\usepackage{tikz}
\usepackage{verbatim}
\usepackage{afterpage}
\usepackage{glossaries}

\usepackage[headsepline]{scrlayer-scrpage}
\clearpairofpagestyles
\automark[subsection]{section}

\usetikzlibrary{arrows,calc}
\def\real{ \mathbb{R} }
\def\nat{ \mathbb{N} }
\newcommand{\bitem}{\begin{itemize}}
\newcommand{\eitem}{\end{itemize}}

\def\real{ \mathbb{R} }
\def\nat{ \mathbb{N} }

\newcommand{\lrbrace}[1]{\left\{ #1 \right\}}

\tikzset{
>=stealth',
help lines/.style={dashed, thick},
axis/.style={<->},
important line/.style={thick},
connection/.style={thick, dotted},
}

\theoremstyle{plain}
\newtheorem{theorem}{Theorem}[section]

\newtheorem{defi}[theorem]{Definition}

\theoremstyle{definition}

\newcounter{ArasCounter}

\newcounter{GittaCounter}

\newcounter{HolgerCounter}

\makeatletter
\g@addto@macro{\thm@space@setup}{\thm@headpunct{}}
\makeatother

\numberwithin{equation}{section}
\allowdisplaybreaks

\begin{document}

\newpage
\setcounter{page}{1}
\clearpairofpagestyles
\ohead{\pagemark}
\ihead{\headmark}

\title{Reliable AI: Does the Next Generation Require Quantum Computing?}

\author{Aras Bacho\footnotemark[2], \ 
	Holger Boche\footnotemark[1] \footnotemark[3] \footnotemark[5], \
  Gitta Kutyniok\footnotemark[2] \footnotemark[4] , 
}
\date{}
\maketitle

\footnotetext[2]{Department of Mathematics, Ludwig-Maximilians-Universität München, Germany}
\footnotetext[1]{Institute of Theoretical Information Technology, TUM School of Computation,
Information and Technology, Technical University of Munich, Germany}
\footnotetext[3]{Munich Center for Quantum Science and Technology (MCQST), Munich, Germany}
\footnotetext[5]{Munich Quantum Valley (MQV), Munich, Germany}
\footnotetext[4]{Munich Center for Machine Learning (MCML), Munich, Germany}

\begin{abstract} 

In this survey, we aim to explore the fundamental question of whether the next generation of artificial intelligence requires quantum computing. Artificial intelligence is increasingly playing a crucial role in many aspects of our daily lives and is central to the fourth industrial revolution. It is therefore imperative that artificial intelligence is reliable and trustworthy. However, there are still many issues with reliability of artificial intelligence, such as privacy, responsibility, safety, and security, in areas such as autonomous driving, healthcare, robotics, and others. These problems can have various causes, including insufficient data, biases, and robustness problems, as well as fundamental issues such as computability problems on digital hardware. The cause of these computability problems is rooted in the fact that digital hardware is based on the computing model of the Turing machine, which is inherently discrete. Notably, our findings demonstrate that digital hardware is inherently constrained in solving problems about optimization, deep learning, or differential equations. Therefore, these limitations carry substantial implications for the field of artificial intelligence, in particular for machine learning. Furthermore, although it is well known that the quantum computer shows a quantum advantage for certain classes of problems, our findings establish that some of these limitations persist when employing quantum computing models based on the quantum circuit or the quantum Turing machine paradigm. In contrast, analog computing models, such as the Blum-Shub-Smale machine, exhibit the potential to surmount these limitations.

\end{abstract}
\vspace*{1em} \textbf{Keywords}  Analog Computer  $ \cdot $ Artificial Intelligence$ \cdot $ Computability $ \cdot $ Computational Complexity $ \cdot $ Deep Learning $ \cdot $ Digital Hardware  $ \cdot $ Quantum Computer $ \cdot $ Reliability of AI $ \cdot $ Turing machine    

\section{Introduction}

The second half of the 20th century saw widespread digitization, which marked the beginning of the third industrial revolution. This digital transformation has driven considerable advancements in science, technology, and productivity across the globe, turning our world into a futuristic landscape. Presently, we find ourselves at the onset of the fourth industrial revolution, which emphasizes automation and interconnectivity, with artificial intelligence playing a pivotal role.

However, as we increasingly depend on ubiquitous technology and our ever-growing knowledge, we encounter numerous challenges and limitations, including the reliability of artificial intelligence and the constraints of digital computers. Broadly speaking, a reliable artificial intelligence system consistently generates accurate, effective, and reliable outcomes. Essential features of such a system encompass high accuracy, robustness, transparency, fairness, privacy, and security.

However, reliability concerns with artificial intelligence (AI) have emerged in various applications, including autonomous driving, facial recognition, healthcare, and robotics. These issues stem from factors such as inadequate data, biases, lack of transparency, security vulnerabilities, and overfitting.

In addition, there are also fundamental problems with digital hardware that cause reliability issues, which may manifest in both physical and theoretical forms. The former, for instance, include the problem of high algorithmic complexity in many algorithms, resulting in an overburdening of the Central Processing Unit. The latter, on the other hand, encompasses the issue of empirical efficiency in many algorithms, particularly in the realm of artificial intelligence, or the potential for unreliable outcomes when the underlying problem is examined at the level of computing models such as the Turing machine.

The Turing machine, conceived by Alan Turing in 1936, serves as a conceptual model to formally encapsulate the notions of algorithms and computability. This theoretical machine, which underpins contemporary computing, comprises a tape partitioned into cells, a head capable of reading and writing on the tape, and a collection of rules that guide the head's operations and mobility based on the symbols it encounters. The Turing machine laid the groundwork for formally understanding algorithms and computability. Subsequently, in 1948, Claude Shannon-often regarded as the pioneer of information theory, devised a mathematical structure to interpret and measure information-greatly influencing the evolution of digital communication and computer science.

Digital computers, which encompass present-day multipurpose computers and digital processors, are ubiquitous in today's world, driving a vast range of technologies from smartphones to supercomputers. However, despite the vast array of computational power available, there are, as mentioned before, fundamental limitations to what can be achieved with digital hardware. 
These limitations necessitate the development of the theoretical foundations for a new type of computing, known as Post-Turing and Post-Shannon computing. Post-Turing computing refers to computing approaches that go beyond the limitations of digital computing as defined by the Turing machine model. Analog and quantum computers have the potential to perform certain types of computations that are not possible or are highly inefficient on digital computers. By exploring the limits of digital computing and identifying specific problems that can benefit from analog or quantum computing, researchers in the field of post-Turing computing can help pave the way for the development of new, more powerful computing technologies. Similarly, Post-Shannon computing refers to computing approaches that surpass the constraints of information processing as established by the Shannon information theory. However, in this survey, we will not delve further into the Post-Shannon theory.

This survey aims to examine the constraints of digital computers rooted in the Turing machine paradigm and to delve into how these restrictions affect the ability to solve specific problem classes, such as partial differential equations or optimization, ultimately influencing the reliability of artificial intelligence. Our research uncovers that many obstacles inherent in the classical Turing machine also persist in quantum computing models like the quantum circuit and quantum Turing machine. Nevertheless, we also provide evidence that analog computing models, such as the Blum-Shub-Smale computation model, can surmount non-computability concerns for certain problem categories. This highlights the promise of real number signal processing hardware, which hinges on the accurate representation of real numbers in overcoming these limitations and propelling the domain of artificial intelligence forward.

\subsection{Outline}
The remainder of the paper is structured in the following way. In Section \ref{se:AI}, we introduce the concept of artificial intelligence and provide a concise overview of deep learning. Furthermore, we explore the question of what criteria an AI system must meet to be reliable and trustworthy, and we discuss the current limitations and challenges of artificial intelligence and identify areas where these limitations are currently problematic. 

In Section \ref{se:Towards pc DL}, we examine the question of provable correctness of deep learning concerning computability on a Turing machine. In Section \ref{se:Foundational ao Computability}, we introduce the basic concepts of computability and complexity theory. Specifically, we introduce the notion of Turing computability and Banach-Mazur computability, as well as notions from computational complexity theory. 

In the main Section \ref{se:limitations}, we wish to discuss the limitations of digital hardware, for both classical computers and quantum computers. In particular, we present negative computability results for a large class of problems in optimization and inverse problems. Furthermore, we present complexity results for differential equations. 

In Section \ref{se:QuantumAnalog}, we discuss the possibilities and limitations of quantum and analog computers that are based on particular computing models. Furthermore, we discuss computability problems and the computational complexity of quantum computers that are based on a quantum Turing machine or quantum circuit with a particular focus on quantum machine learning. Finally, we present positive computability results for the analog computing model of a Blum-Shub-Smale machine. 

The paper closes with some concluding remarks in Section \ref{se:outlook}.

\section{Artificial Intelligence} \label{se:AI}
Artificial intelligence encompasses the area of computer science and engineering dedicated to creating intelligent agents. These agents are computer programs capable of executing tasks that usually demand human intelligence, such as perception, reasoning, learning, and natural language comprehension. AI systems are engineered to process vast quantities of data, identify patterns, and make predictions or decisions based on that information, often employing machine learning algorithms. AI has applications across a wide range of fields, including computer vision, game playing, natural language processing, and robotics, among others.

The evolution of AI has been propelled by advancements in computer hardware, software, and data, as well as by breakthroughs in cognitive psychology and neuroscience. These advancements have led to significant improvements in AI technology. As AI continues to progress, it is anticipated to significantly impact various aspects of society, such as healthcare, transportation, education, and entertainment.

AI systems can be categorized based on their degree of autonomy and their capacity to learn and adapt. For instance, rule-based systems rely on predetermined rules for decision-making, while machine learning algorithms can learn from data and enhance their performance over time. Machine learning embodies the long-standing concept that machines can learn to tackle specific problems independently, given access to the appropriate data.

Utilizing sophisticated mathematical and statistical tools, machine learning aims to deliver computing power capable of autonomously addressing intellectual tasks that have conventionally been tackled by humans. This automation of complex tasks has sparked considerable interest in diverse fields, such as automotive, cybersecurity, education, entertainment, finance, healthcare,  manufacturing, and marketing.

The ongoing development of AI and machine learning will continue to transform our society in unforeseeable ways. As these technologies advance, it is crucial to remain aware of the potential benefits and challenges they present and to ensure that their deployment aligns with ethical principles and societal values.

\subsection{Deep Learning} \label{se:deep.learning}
One instance of machine learning paradigms recently enjoying a great deal of attention is the so-called deep learning. The popularity of deep learning is justified by the remarkable success of its usage, leading to breakthroughs in diverse applications \cite{Jumper2021HighlyAP, Brown20GPT3}. The inspiration for deep learning comes from biology, as it utilizes an architecture based on the human brain, called an (artificial) neural network. A neural network consists of a collection of connected units or nodes subdivided into several layers allowing the network to learn several abstraction levels of the input signal \cite{Berner21MMDL}.

Formally, a \textit{(feed-forward) neural network} is defined as the mapping $\Phi:\mathbb{R}^{N_{0}}\rightarrow\mathbb{R}^{N_{L}}$ having the form:
\begin{equation}
\label{Eq:FeedForward}
    \Phi(x)=W_{L}(\rho(W_{L-1}(\rho(\cdots\rho(W_{1}(x)))))),
\end{equation}
where $L\in\nat$, $L\geq 2$, $\mathbf{N}:=(N_{L},N_{L-1},\ldots,N_{1},N_{0})\in\nat^{L+1}$, where $W_{j}:\real^{N_{j-1}}\rightarrow \real^{N_{j}}$ is affin, for all $j\in [L]$, and where $\rho:\real\rightarrow\real$ is a non-linear function acting component-wise on a vector. Neural networks having the form \eqref{Eq:FeedForward} are also called $L$-layer neural networks with $\mathbf{N}$ neurons. The set of such networks is denoted by $\mathcal{NN}_{\mathbf{N},L}$.
Deep learning is then called the process that aims to 'learn' parameters $W_{j}:\real^{N_{j-1}}\rightarrow \real^{N_{j}}$ such that the neural network \eqref{Eq:FeedForward} approximates the ground truth function $f$ of input-output pairs $\lbrace (x_i,y_i) \rbrace_{i=1}^{m} $ with $f(x_i)=y_i$ possibly corrupted with noise. This is accomplished by defining a loss function $\mathcal{L}:\mathbb{R}^m\times \mathbb{R}^m \rightarrow \mathbb{R}$ and seeking a minimizer of the empirical loss function 
\begin{equation}
\min_{\Phi} \sum_{i=1}^m \mathcal{L}(f(x_i),y_i),
\end{equation} aiming for a function $\Phi$ to approximate $f$, i.e., $\Phi \approx f$. A typical choice for the cost functional is the square function. The minimization is typically done by gradient decent methods such as stochastic gradient decent. For classification tasks, one usually discretizes the output of $\Phi$ by means of a discrete-valued function $f:\real^{N_{L}}\rightarrow\lrbrace{0,1}^{D}$, where $D\in\nat$, yielding a classificator $\tilde{\Phi}=f\circ\Phi:\real^{N_{0}}\rightarrow\lrbrace{0,1}^{D}$.

Even with their achievements, the core characteristics of deep neural networks remain incompletely comprehended, prompting a rigorous investigation in recent times. Common inquiries in this regard include:

\begin{itemize}
    \item What is the theoretical basis for deep learning's effectiveness?
    \item How can the choices made by a deep neural network be interpreted?
    \item Are there any trade-offs associated with deep learning's success?
    \item What are the boundaries of deep learning's capabilities?
    \item Can deep learning consistently yield reliable outcomes?
\end{itemize}


\subsection{Reliability of Artificial Intelligence} \label{se:Reliability}
Trustworthy artificial intelligence pertains to AI systems that continually execute tasks with precision, efficiency, and safety, yielding credible and consistent outcomes. Such systems must demonstrate resilience against a range of obstacles, including unpredictable or dynamic environments, noisy information, and potential malicious attacks. Additionally, a reliable AI system ought to be transparent and comprehensible, providing users insight into its decision-making mechanisms and empowering developers to pinpoint and rectify any possible issues or biases.

To achieve this level of reliability, an AI system should consider the following aspects as a general framework \cite{GMRTGP18SMEBB,Abbas2022SafetySA,Guidotti2018ASO,sculley2015hidden,Bottou2010LargeScaleML,BaHaNa19FMLO,Jobin2019TheGL,Arrieta19EAI,GoBenCou16DL}:

\begin{enumerate}

\item \textbf{Accuracy:} Trustworthy AI systems must consistently deliver accurate and precise outcomes when addressing complex tasks. Their performance should remain at a high level, with errors minimized to ensure the reliability of the generated results.

\item \textbf{Generalizability:} A reliable AI should be able to effectively apply its learned knowledge from training data to new, unseen data or situations. It's a critical feature ensuring that an AI model performs accurately and consistently across various situations.

\item \textbf{Robustness:} Reliable AI systems should exhibit resilience against a variety of challenges, including changing environments, noisy or incomplete data, and adversarial attacks. They must be designed to manage uncertainties and unexpected inputs without compromising performance or leading to unintended consequences.

\item \textbf{Transparency:} Establishing trust in AI systems necessitates transparency in their decision-making processes. Reliable AI should offer clear justifications for predictions or recommendations, enabling users to comprehend and validate their actions.

\item \textbf{Interpretability:} Reliable AI should provide insight into its internal functions, allowing developers and users to grasp the reasoning behind its decisions. This understanding facilitates the identification and rectification of potential issues, biases, or undesirable behaviors.

\item \textbf{Fairness:} A reliable AI system must treat various user groups equitably and avoid discrimination or bias. It should be designed and tested to ensure its decisions do not unintentionally favor one group over another due to biases in the training data or algorithms.


\item \textbf{Security:} A reliable AI system must be safeguarded against potential threats, both internal and external. This security entails protection against unauthorized access, data breaches, and manipulation of the AI's decision-making process.

\item \textbf{Safety:} Reliable AI systems must prioritize user and environmental safety during their operation. They should be designed to minimize risks, prevent unintended harmful consequences, and ensure that any potential hazards are identified and mitigated. By focusing on safety, AI systems can be trusted to operate within established guidelines and ethical standards, ensuring a secure experience for all users and stakeholders involved. 

\item \textbf{Privacy:} Reliable AI must respect and preserve user privacy, ensuring that sensitive information is managed securely and in accordance with relevant regulations and ethical guidelines.

\item \textbf{Scalability:} For AI systems to be considered reliable, they should efficiently scale, accommodating increased workloads and data volumes without sacrificing performance or accuracy.


\end{enumerate}
We make no claim to the completeness of this list, and one could further include important attributes such as accountability, maintainability, ethical compliance, usability, sustainability, among others. Another important aspect is that the properties are not completely independent of one another. For instance, safety also entails security.

Consequently, reliable artificial intelligence involves an extensive array of features that contribute to the trustworthiness, safety, and efficiency of AI systems. For critical applications such as autonomous driving, robotics, or medicine, these characteristics of a reliable AI are indispensable, and incorporating these attributes in AI systems will result in broader acceptance and a beneficial influence across diverse industries and applications.

\subsection{Limitations of Artificial Intelligence} \label{se:Limitations of AI}

A significant challenge currently faced by artificial intelligence, especially in the realm of deep learning, is the absence of dependability. Reliability issues can be observed in various domains and applications such as healthcare, autonomous driving, criminal justice, social media, natural language processing, and computer vision, among others. This issue is critical due to several contributing factors \cite{Goodfellow2014ExplainingAH,Domingos2012AFU,gilpin2018explaining,FHCJM19EFSE,BaHaNa19FMLO,Stone2016ArtificialIA,FerrariDacrema2019AreWR,sculley2015hidden,Jobin2019TheGL,Strubell2019EnergyAP,Gorecky2014HumanmachineinteractionIT,Amershi2019GuidelinesFH}:

\begin{enumerate}

\item \textbf{Data quality and biases:} The performance of AI systems, especially machine learning models, heavily depends on the quality of training data. If the data is noisy, incomplete, or biased, it may lead to unreliable or unfair outcomes.

\item \textbf{Interpretability and explainability:} A large number of AI models, such as deep learning models, are viewed as "black boxes" due to their intricate internal workings. This makes understanding and explaining their decision-making processes difficult, affecting their trustworthiness and adoption in critical applications.

\item \textbf{Generalization and overfitting:} AI models should generalize effectively from training data to new, unseen data. However, overfitting the training data may result in poor performance on new data, leading to unreliable predictions.

\item \textbf{Robustness and adversarial attacks:} AI systems can be susceptible to adversarial attacks or manipulations, where small, intentional perturbations in input data cause incorrect or deceptive outcomes. Ensuring robustness against such attacks is essential for AI systems' reliability.

\item \textbf{Ethical concerns and fairness:} AI systems may unintentionally reinforce or worsen existing biases and stereotypes found in training data, resulting in unfair treatment for certain groups or individuals. Addressing these ethical issues and promoting fairness in AI systems is crucial for their reliability.

\item \textbf{Complex interactions in real-world environments:} AI systems frequently operate in intricate, dynamic settings where they interact with humans and other AI systems. Managing these complexities and uncertainties while maintaining reliable performance is a considerable challenge.

\item \textbf{Lack of standardized evaluation metrics:} Standardized evaluation metrics and benchmarks for measuring the reliability and performance of AI systems across various domains and applications are often absent. This lack of standardization complicates the assessment and comparison of different AI systems' reliability.

\item \textbf{Rapid development and deployment:} The accelerated development and deployment of AI technologies may lead to inadequate testing, validation, and evaluation of AI systems before their real-world implementation. This could result in systems that are not sufficiently reliable or robust.

\item \textbf{Legal and regulatory challenges:} Current legal and regulatory frameworks for AI are still developing, potentially leaving gaps in addressing AI systems' reliability and safety. Ensuring the reliability of AI systems within these contexts remains an ongoing challenge.

\item \textbf{Computational resources:} Developing and deploying AI models, particularly large-scale deep learning models, can demand considerable computational resources, creating a barrier for smaller organizations and researchers.

\end{enumerate}

This becomes an even more pressing problem, considering the AI Act of the European Union \cite{WebsiteEuCom}, which enforces reliability certificates for AI technology to get approved. Another initiative is the so-called 'Hiroshima AI Process,' in which the G7 countries discuss, among other topics, the advantages and risks of generative AI and how to promote AI in general, see, e.g., \cite{Yao23G7GA}. 

Therefore, understanding the described reliability issues is of tremendous importance to enable trustworthy AI applications in the future. One well-known failure in this regard is the instability phenomenon of deep neural networks, which is one of the most prominent disadvantages of this learning paradigm \cite{Fawzi2017,Nguyen2015,Eykholt2018,Antun2020}:  
An arbitrarily small (invisible for humans) perturbation of the input can significantly reduce the performance of a neural network predictor. One prototypical example of this deep learning vulnerability is given in \cite{Moosavi-Dezfooli16}, where a deep neural network classifies a slightly perturbed whale image as a turtle image, although the used architecture classifies the original image correctly.

\subsection{Towards Provable Correctness of Deep Learning} \label{se:Towards pc DL}
A theoretical analysis of a numerical method might not always consider the aspect of computability, which relates to the method's effectiveness when implemented on a digital computer, as discussed in Section \ref{se:limitations}. Both positive and negative answers to this question carry practical implications. If the method can be effectively implemented, it becomes useful in real-world situations where digital technologies are prevalent. Conversely, if the method cannot be effectively implemented, this insight can help designers identify potential drawbacks and necessitate a change in approach.

Recognizing the limitations of digital computations is vital when examining the properties of deep learning, both during training and deployment. This is because the characteristics of the underlying computing device can significantly influence the performance of a deep learning system. By identifying these limits, we can determine which tasks deep learning systems can reliably perform on digital machines and assess the potential for future enhancements. Addressing these fundamental questions is crucial before delving into the instability phenomenon mentioned earlier. Moreover, it is essential to ascertain whether this inadequacy stems from inherent mathematical properties or the current implementations of deep learning systems. Thus, evaluating the boundaries of digital computations is key to obtaining insightful answers to these questions.

\section{Foundational Aspect of Computability} \label{se:Foundational ao Computability}
In this section, we wish to introduce notions from computability and computational complexity theory. 
\subsection{Computability Theory}\label{se:computabiliy}
One of the main limitations of digital hardware is its inability to perform exact computations on continuous quantities such as real numbers. Digital computers use binary representations of numbers, which are inherently discrete and finite, making it impossible to represent real numbers exactly. As a result, digital computers can only approximate real numbers, which leads to inaccuracies and errors in computations. 

The \textit{computability theory}, also referred to as \textit{recursion theory} or the \textit{theory of computability}, is a prominent field of study within mathematical logic that delves into the fundamental limitations and capabilities of computational processes. It explores which problems can be solved algorithmically, and which problems cannot be solved by any computational procedure. The theory examines the concept of an algorithm, how algorithms can be represented and manipulated, and the properties of algorithms such as their efficiency, correctness, and the types of problems they can solve. Additionally, computability theory provides a foundation for the study of complexity theory and the design and analysis of algorithms. Modern computability theory starts with the seminal work of Turing \cite{Turing1936} introducing the concept of a Turing machine constituting a model of computing unit. The bridge between the abstract implementation-technical concept of a Turing machine and the mathematical analysis is provided by the concept of the so-called (partially) recursive functions previously introduced by G\"odel \cite{Goedel1931} giving rise to the formal notion of effective basic mathematical objects such as computable numbers and functions. 
\begin{defi}(Computable number) \label{def:TM}
    A number $t\in \mathbb{R}$ is said to be \emph{computable}, if there exists a Turing machine TM with input $n \in \mathbb{N} $ and output $\gamma(n)=TM(n)\in \mathbb{Q} $, such that
\begin{equation}\label{computable.number}
    \vert t-\gamma(n)\vert\leq 2^{-n}, \quad \text{for all }n\in \mathbb{N}.
\end{equation}
In this case, we say that $\gamma(n)$ \emph{binary converges to} $t$, and we write
$\mathbb{R}_c \subsetneq \mathbb{R}$ for the set of all computable real numbers.
\end{defi}
We note that the set $\mathbb{R}_c$ of computable numbers are dense in $\mathbb{R}$ and closed under addition, subtraction, multiplication, and division by numbers in $\mathbb{R}_c\backslash\lbrace 0\rbrace$. The Turing machine TM defines the limit of real numbers that are computable in the sense of Definition \ref{def:TM}. However, it does not specify the number of iterations (i.e. computation time) that are required for the Turing machine to calculate $\gamma(n)$ for a given input $t\in \mathbb{R}_c$. Since we are interested in determining the computational complexity of certain classes of problems, the next definition quantifies the number of iterations that are required to approximate $t\in \mathbb{R}_c$ as $n$ increases. 
\begin{defi}(Polynomial-time computable number)
    Let $t\in \mathbb{R}_c$ be a computable number. We say that the \emph{computational complexity of $t$ is bounded by a function} $q:\mathbb{N}\rightarrow \mathbb{N}$, if there exists a Turing machine TM such that \eqref{computable.number} is satisfied after at most $q(n)$
iterations. The number $t\in \mathbb{R}_c$ is said to be \emph{polynomial-time computable} if its computational complexity is bounded by a polynomial $q$.
\end{defi}
To define computable functions, we employ the concept of a function-oracle Turing machine. An oracle Turing machine is a Turing machine that is equipped with a function oracle that calculates the value of $\gamma$ from Definition \ref{def:TM} in a single step. However, the previous definition suggests that the calculation of $\gamma$ might take some computational time. This concept allows quantifying the computational complexity of a function distinct from the computational complexity of the input data, which requires a Turing machine itself to be approximated. 

\begin{defi}(Computable function) \label{def:comp.func}
    Let $x:[a,b]\rightarrow \mathbb{R}$ be a real function. Then $f$ is said to be \emph{computable on the interval}  $[a, b] \subset \mathbb{R}$, if there exists a function oracle Turing machine TM, such that for each $t \in [a, b]$ and each $\gamma$ that binary converges to $t$, the
function $\tilde{x}(n) = TM_\gamma (n)$ computed by TM with oracle $\gamma$ binary converges to $x(t)$, i.e. if
\begin{equation}\label{computable.function}
    \vert x(t)-\tilde{x}(n)\vert\leq 2^{-n}, \quad \text{for all }n\in \mathbb{N}.
\end{equation}
\end{defi}

With the definition of a computable function, we can define a polynomial-time computable function.
\begin{defi}(Polynomial-time computable function)\label{def:ptc.oracle}
   Let $x : [a, b] \rightarrow \mathbb{R}$ be a computable function. We say that the \emph{complexity of $x$ is bounded by a function} $q : \mathbb{N}\rightarrow \mathbb{N}$, if there exists a function–oracle Turing machine TM, which computes $x$ such that for all $\gamma$, that binary converges to a real number $t \in [a, b]$, and
for all $n\in \mathbb{N}, TM_\gamma (n)$ satisfies
\begin{equation}\label{computable.function.poly}
    \vert x(t)-TM_\gamma(n)\vert\leq 2^{-n}
\end{equation}
after a computation time of at most $q(n)$. The function $x :
[0, 1] \rightarrow \mathbb{R}$ is said to be \emph{polynomial-time computable}, if its complexity is bounded by a polynomial $q$.
\end{defi}

Based on these definitions, complexity can be assigned to the solution of problems, or it can even be shown that the solutions to problems are not Turing-computable or even Banach-Mazur computable, which is a weaker notion of computability. To formally introduce this concept of computability, we will present an equivalent definition of computable numbers. Specifically, a \textit{computable (real) number} is defined as a number approximable by a computable sequence of rational numbers $(r_{n})_{n\in\mathbb{N}}$ having the form $r_{n}=(-1)^{s(n)}\frac{a(n)}{b(n)}$, $n\in\mathbb{N}$, where $a,b,s:\mathbb{N}\rightarrow\mathbb{N}$ are the so-called recursive functions with $b(n)\neq 0$ for all $n\in\mathbb{N}$. Coupled with this notion of a computable number is the notion of a \textit{(Banach–Mazur) computable function} $f:\real\rightarrow\real$, defined as a function mapping any computable sequence of real numbers to a computable sequence of real numbers.

Now, according to the Church–Turing thesis, a function on the natural numbers can be calculated by an effective method if and only if it is Turing-computable. 
Furthermore, it is a well-known fact that (by omitting the finite memory assumption) the abstract model of several widely used powerful programming languages, such as C++ and Java, can compute every Turing-computable function. Therefore, the theoretical aspects of computability are essential for the practical usage and implementation of signal, information, and data processing methods such as deep neural networks, as the universal Turing machine constitutes the digital computer's ultimate limit.

\subsection{Computational Complexity Theory}\label{se:complexity}

In the following, we will introduce certain complexity classes that capture and define the complexity of evaluating a given function reasonably. In particular, we introduce complexity classes for decision, counting, and function problems. Then, with the definition of the computability of a function on a dyadic grid, we will be able to formulate the problem of evaluating a function as a counting problem. 
\subsubsection{Decision Problems and the Classes P and NP}
To investigate the computational complexity of solutions to partial differential equations, we need to introduce appropriate complexity classes. The best-known complexity classes are $P$ and $NP$ which consist of \textit{decision problems}. Decision problems are problems that require a  `yes' or `no' answer. The class $P$ consists of all decision problems that are solvable in polynomial time by a deterministic Turing machine meaning that the computational complexity grows polynomially in the input size. The class $NP$ consists of all decision problems for which a given answer can be verified in polynomial time by a deterministic Turing machine. It is obvious that $P\subset NP$. However, the question of whether $P= NP$ or $P\subseteq NP$ remains a major open problem in theoretical information theory and belongs to the famous Millennium Prize Problems.
\subsubsection{Counting Problems and the Classes $\#P$ and $\#P_1$} 
Another complexity class is given by the set of \textit{counting problems}, which does not ask whether a given problem in $NP$ has a solution but enumerates the number of solutions. Clearly, the class $NP$ is contained in $\#P$. 
For a more formal definition of $\#P$, let $\lbrace 0,1\rbrace^n$ be the set of all words
of length $n\in \mathbb{N}$ in the alphabet $\Sigma=\lbrace 0,1\rbrace$ consisting of 0 and 1, and let $\Sigma^*$ denote the set of all finite words in the alphabet $\Sigma$. For a given string $x\in \Sigma^*$, we denote the length of $x$ with $len[x]$. Then, a \emph{function $f : \Sigma^* \rightarrow \mathbb{N}$ is in $\#P$}, if there exists a polynomial $p:\mathbb{N}\rightarrow \mathbb{N}$ and a polynomial-time Turing machine $M$, so that for every string $x\in \Sigma^*$, we have 
    \begin{equation}
        f(x)=\left \vert \left \lbrace  y\in \Sigma^{p(len[x])}: M(x,y)=1 = \text{`Yes'} \right \rbrace \right \vert.
    \end{equation}
Here $f(x)$ denotes the number of accepting paths (or certificates) for the input $x$. Similarly, we define the subclass of counting problems by restricting the set of counting problems to functions of the form $f:\lbrace 0\rbrace^*\rightarrow \mathbb{N}$,  where $\lbrace 0\rbrace^* =\lbrace \lbrace 0\rbrace, \{0, 0\}, \{0, 0, 0\}, \dots \}$. We denote this class with $\#P_1$. More formally, a function $f : \lbrace 0\rbrace^* \rightarrow \mathbb{N}$ is in $\#P_1$, if there exists a polynomial $p:\mathbb{N}\rightarrow \mathbb{N}$ and a polynomial-time Turing machine $M$, so that for every string $x\in \lbrace 0\rbrace^*$, we have 
\begin{equation}
        f(x)=\left \vert \left \lbrace  y\in \lbrace 0\rbrace^{p(len[x])}: M(x,y)=1 \right \rbrace \right \vert.
\end{equation} 
It is obvious that $\#P_1\subset \#P$. A very prominent and important problem that belongs to $\#P$ is the task of calculating the permanent of a matrix with entries consisting of $0$'s and $1$'s which is related to the Boson sampling problem in quantum computing \cite{Aaronson2010TheCC}. This problem is even $\#P$-complete meaning that there exists a deterministic Turing machine that can use the $\#P$-complete problem to simulate any other problem in the complexity class $\#P$ in polynomial time. Hence, if there exists a polynomial-time Turing machine that solves a problem in $\#P$-complete, then any other problem in the same complexity class can be solved in polynomial time. In other words, the $\#P$-complete problems are the hardest among all problems in $\#P$. Similarly, one can define the property of being complete for other complexity classes. Another problem in $\#P$ is the graph coloring problem which asks for the number of admissible colorings using $k\in  \mathbb{N}$ colors for a particular graph. 
\subsubsection{Function Problems and the Classes $FP$ and $FP_1$}
Similar to the class of decision problems that can be solved in polynomial time, we can define the class of counting problems that can be solved by a function oracle Turing machine in polynomial time denoted by $FP$ and $FP_1$ which are also called \textit{function problems}. Formally, the classes are defined as follows:
\begin{itemize}
    \item A function $f : \lbrace 0,1\rbrace^* \rightarrow \mathbb{N}$ belongs to $FP$, if it can be computed by a deterministic Turing machine in polynomial time.
     \item A function $f : \lbrace 0\rbrace^* \rightarrow \mathbb{N}$ belongs to $FP_1$, if it can be computed by a deterministic Turing machine in polynomial time.
\end{itemize}
By the definition of the classes $FP$ and $FP_1$, it is evident that $FP\subset \#P$ and $FP_1\subset \#P_1$. Similar to the $P$ vs. $NP$ problem, the question arises whether $FP= \#P$ and $FP_1= \#P_1$. Furthermore, it can be shown that $P=NP$ implies $FP= \#P$ and $FP_1= \#P_1$. As the equality $P=NP$ would have immense consequences in many fields, it is widely assumed that $FP \subsetneq \#P$ and $FP_1 \subsetneq \#P_1$.
It is easy to see that we can identify the set of finite strings $\Sigma^*$ over $\Sigma$ with the set of dyadic decimals $D_1\cap[0,1]$ in a canonical way: to a sequence $d_1d_2d_3\dots d_n$, we uniquely associate the binary number $d=0. d_1d_2d_3\dots d_n \in D_1\cap[0,1]$.

\section{Limitations of Computability on Digitial Hardware} \label{se:limitations}
In this section, we aim to introduce several findings which demonstrate that numerous classes of problems are provably intractable when utilizing digital hardware. These include inverse problems, initial and boundary value problems in the theory of ordinary and partial differential equations as well as optimization problems such as optimizing a neural network in artificial intelligence, solving the lattice problem in post-quantum cryptography, or maximizing the information flow in information theory. In particular, these results show how the limitations of digital hardware impact the reliability of artificial intelligence. We start by presenting these results for optimization problems. 

\subsection{Optimization Problems} \label{se:optimization}
In the study by \cite{Lee2023ComputabilityOO}, the authors conduct a systematic analysis of the constraints imposed by digital hardware on solving optimization problems from a computability perspective. They explore whether the digital nature of current hardware restricts the computability of solutions to optimization problems, which often have continuous solutions. To address this question, they develop an extensive theory of non-computability or non-approximability for optimization problems within the broader context of non-Banach-Mazur computability.

Their primary finding, as presented in \cite[Theorem 1]{Lee2023ComputabilityOO}, reveals that for a significant class of optimization problems, the computation or approximation of the optimizer, represented by the function $G: Y \rightarrow X$, is not possible in a computable manner. However, they also show that this result does not necessarily mean that the optimal value, a function $\Phi: Y \rightarrow \mathbb{R}$, or an approximation of it, is non-computable. To further clarify the implications of their findings, they present a variety of optimization problems where the optimizer is non-computable and non-approximable, yet the corresponding optimal value is computable. Examples of these problems include convex optimization challenges such as neural networks, portfolio optimization, information theory, optimal transport, cryptography, and linear programming.

Their findings hold significant implications for a wide array of applications, including quantum machine learning, thereby contributing to a deeper understanding of the limitations classical digital hardware and quantum computers impose on solving optimization problems. In the following sections, we will discuss essential applications of their main result, encompassing neural networks, information theory, cryptography, and linear programming.

\subsubsection{Neural Networks} \label{se:neuralnetwork}
Now, we present a negative result on the reliability of deep learning and recall that in deep learning one aims to find parameters $W_{j}:\real^{N_{j-1}}\rightarrow \real^{N_{j}}$ of the neural network defined in \eqref{Eq:FeedForward} 
that approximates the ground truth function $\Psi$ of input-output pairs $\lbrace (x_i,y_i) \rbrace_{i=1}^{m}$ with $\Psi(x_i)=y_i$ by minimizing the empirical loss function 
\begin{equation}
\min_{\Phi} \sum_{i=1}^m C(\Phi(x_i),y_i).
\end{equation} Next notice that \cite[Theorem 2]{Lee2023ComputabilityOO} states that the resolvent operator $G: \mathbb{R}_c ^m\times \mathbb{R}_c^m\rightarrow \mathcal{NN}_{\mathbf{N},L}$ which maps computable observations to a minimizing shallow neural network is, in general, not Banach-Mazur computable. This implies that the optimal weights are, in general, not computable. As a consequence, the result of any minimization technique including (stochastic) gradient descent is not reliable. This negates the question of whether deep learning produces reliable results on digital hardware.\,Although the result was demonstrated only for shallow neural networks, the authors claim that this result holds for general neural networks and that this result can be an indication of why instabilities are observed in some applications \cite{Gottschling2020TheTK}.

\subsubsection{Information Theory}

Information theory is a field within mathematics and computer science that focuses on the quantification, storage, and communication of information. Developed by Claude Shannon in the late 1940s and early 1950s, information theory has numerous practical applications, such as coding theory, data compression, cryptography, and digital signal processing. It has also been employed to examine complex systems like biological and social networks, as well as to analyze the behavior of financial markets and other large-scale systems.

The primary objective of information theory is to study the fundamental properties of communication systems, including the capacity to transmit information over a channel, error minimization during information transmission, and information compression to decrease storage and transmission costs. The central concept in information theory is the "bit," which serves as the basic unit of information. A bit can assume one of two possible values, such as 0 or 1, and can be used to represent a binary decision or a binary digit in a computer system. The amount of information conveyed by a sequence of bits depends on the number of possible sequences and the probability distribution of those sequences.

One can model the information transmitter as a discrete random variable $X$ over a probability space $\mathcal{X}$ and probability density function $P_{X}$, and the information receiver as a discrete random variable $Y$ over a probability space $\mathcal{Y}$ and probability density function $P_{Y}$. The mutual information (transformation) is then given by

\begin{equation}
I(X;Y)=\sum_{x\in \mathcal{X}}\sum_{y\in \mathcal{Y}} P_{(X,Y)}(x,y)\log\left( \frac{P_{(X,Y)}(x,y)}{P_{X}(x)P_{Y}(y)} \right),
\end{equation} where $P_{(X,Y)}$ is the joint probability density function of $(X,Y)$. It quantifies the amount of information that one variable provides about the other variable. The goal is to maximize the mutual information between the transmitter $X$ and the receiver $Y$ so that $Y$ receives as much information as possible. More formally, one aims to maximize the capacity $C$ defined by:
\begin{equation} \label{}
C= \sup_{P_X} I(X,Y)
\end{equation}

where the supremum is taken over all input probability distributions $P_X$. Since for a fixed conditional probability distribution $P_{(Y\vert X)}$, the marginal distribution $P_X$ is completely determined by the joint distribution $P_{(X,Y)}$ through the formula 
\begin{equation}
P_{(X,Y)}(x,y) = P_{(Y\vert X)}(y,x) P_{X}(x).
\end{equation}

Notice that this optimization problem is well-defined. Let $\mathcal{P}(\mathcal{X})$ represent the set of random variables $X$ over probability space $\mathcal{X}$. Now, \cite[Theorem 4]{Lee2023ComputabilityOO}, thereby also generalizing the results in  \cite{BoScPo22CAID}, states that the resolvent operator $G: \mathbb{R}_c^{m\times n}\rightarrow \mathcal{P}(\mathcal{X})$, which maps the object conditional probability distribution $P{(Y\vert X)}$ to the optimal distribution $P_X$ that maximizes the capacity, is not Banach-Mazur computable. 

\subsubsection{Post-Quantum Cryptography - The Lattice Problem}

The lattice problem is a central issue in lattice-based cryptography. In this problem, we are given a lattice in a $n$-dimensional space, and the objective is to find a short vector within the lattice. More specifically, we want to find a vector $v$ that is shorter than a certain bound and is part of the lattice. The lattice problem is considered to be computationally difficult (NP-hard), meaning that there is no known efficient algorithm capable of solving the lattice problem for every instance. This hardness assumption serves as the basis for numerous lattice-based cryptographic schemes, including encryption, digital signatures, and key exchange protocols.

Lattice-based cryptography has gained significant interest due to its security properties and resistance to quantum computer attacks. Moreover, it has been employed in various applications, such as post-quantum cryptography, privacy-preserving machine learning, and secure multiparty computation. In the following, we focus on a specific lattice problem, namely the shortest vector problem. To formally define this problem, let $B=\lbrace b_1,\dots,b_n\rbrace$ be a basis of the vector space $V$ and $\Vert \cdot \Vert_V :V\rightarrow \mathbb{R}$ be a norm on $V$. Additionally, define the lattice

\begin{equation}
\Lambda(B):=\left \lbrace \sum_{i=1}^n \lambda_i b_i \mid \lambda_i\in \mathbb{Z}, i=1,\dots, n \right \rbrace
\end{equation} and the set of bases:
\begin{equation}
\mathcal{B}(B):=\lbrace \beta\subset \Lambda(B)\mid \beta \text{ is a basis of } \mathbb{R}^n\rbrace.
\end{equation}

We use $\lambda(\beta)$ to represent the length of the shortest non-zero vector in the lattice $\Lambda$, that is,

\begin{equation}\label{eq:basis}
\lambda(\beta) = \min_{\beta \in \mathcal{B}(B) }\sum_{b \in \beta} |b|_V.
\end{equation}

Let $\Vert \cdot \Vert_V=\Vert \cdot \Vert_2$ be the $\ell^2$ norm and let $\mathcal{V}$ denote the set of all bases of $\mathbb{R}^n$. Then, \cite[Theorem 6]{Lee2023ComputabilityOO} states that the resolvent operator $G:\mathcal{V}\rightarrow \mathcal{V}$, which maps a basis $B$ to the basis $B$ that satisfies \eqref{eq:basis}, is not Banach-Mazur computable.

Although this result may seem negative, it is positive for cryptographic purposes because it asserts that an attacker cannot generally compute the optimal basis. Given that cryptographic methods that are based on the lattice problem have already been categorized as post-quantum and are therefore believed to be secure against quantum computers, this non-computability result further strengthens this assumption.

\subsubsection{Linear Programming}

Linear programming is an optimization technique in mathematics used to find the best possible solution to a problem with linear constraints and a linear objective function. This method involves identifying a set of decision variables and determining their optimal values to maximize or minimize the objective function while meeting the constraints. In a linear programming problem, the objective function is a linear equation representing the quantity to be maximized or minimized. The constraints are linear equations or inequalities that restrict the values the decision variables can assume.

Linear programming problems can always be expressed as follows: For a given cost vector $c \in \mathbb{R}^m$, find $x\in \mathbb{R}^m$ such that

\begin{equation}
c^\mathsf{T} x\geq \max_{A y \leq b}c^\mathsf{T} y,
\end{equation}

where $b\in \mathbb{R}^{n}$ and $ A\in \mathbb{R}^{m\times n}$, and $A y  \leq b$ represents the set of constraints, understood component-wise.

Linear programming is widely utilized in fields such as economics, engineering, operations research, and management science to address complex issues like resource allocation, production planning, transportation logistics, and investment portfolio management. The simplex algorithm is a common method for solving linear programming problems.

According to \cite[Theorem 7]{Lee2023ComputabilityOO}, the resolvent operator $G: \mathbb{R}_c ^{m\times n}\times \mathbb{R}_c^n \times \mathbb{R}_c^m \mathbb{R}_c^m$, which maps the object $(A,b,c)$ to an optimizer $x$, is not Banach-Mazur computable.

\subsection{Inverse Problems}\label{se:inverse.problems}

Inverse problems are a category of problems where the goal is to identify the input or cause that generated a specific output or effect. In other words, given observed or measured data, an inverse problem aims to discover the underlying parameters, functions, or variables responsible for creating that data. Inverse problems are widespread in various scientific, industrial, and medical applications such as electron microscopy, seismic imaging, magnetic resonance imaging, and computed tomography. For instance, in medical imaging, an inverse problem could involve reconstructing a three-dimensional image of a patient's internal organs from a series of two-dimensional X-ray images. Here, the observed data are the X-ray images, while the underlying parameters or variables are the shape and position of the organs. 

Inverse problems are often challenging to solve as they are ill-posed in the sense of Hadamard, i.e., existence, uniqueness, and stability of solutions are in general not guaranteed: multiple possible solutions can be consistent with the observed data, making it difficult to determine the correct one. Additionally, the observed data may be noisy or incomplete, further complicating the problem. 

The algorithmic solvability of the following finite-dimensional inverse problems on digital hardware was explored in \cite{Boche2022LimitationsOD}:
\begin{center}
\emph{Given noisy measurements $y=Ax+e \in \mathbb{C}^m$ of $Ax \in \mathbb{C}^m$, recover $x\in \mathbb{C}^N$},\end{center} 
where $A\in \mathbb{C}^{m \times N}, m<N$, is the sampling operator or measurements operator, $e\in \mathbb{C}^m$ is a noise vector, $y \in \mathbb{C}^m$ is the vector of measurements and $x \in \mathbb{C}^N$ is the object to recover. The authors showed even that the solution operator which maps $y \in \mathbb{C}^m$ and $A\in \mathbb{C}^{m \times N}$ to the solution of the relaxed inverse problem given by the optimization problem
\begin{equation}
    \min_{x\in \mathbb{C}^N} \Vert x\Vert_{\ell^1} \quad \text{such that } \Vert Ax-y\Vert_{\ell^2}\leq \varepsilon
\end{equation} is not Banach-Mazur computable, where $\varepsilon>0$ denotes the relaxation parameter.  
The study revealed that any method operating on digital hardware is subject to specific constraints when approximating the solution maps of inverse problems. This finding establishes a fundamental limitation on current digital computing devices. The restriction is that any single-valued constraint of the solution mapping of the optimization problem is not Banach-Mazur computable. This implies that there is no reliable general learning algorithm applicable to all inverse problems.

\subsection{Pseudo Inverse} \label{se:pseudoinverse} 

The pseudoinverse is a mathematical concept that extends the idea of the inverse of a matrix. While the inverse of a matrix only exists for square, invertible matrices, the pseudoinverse can be defined for any matrix, irrespective of its size or invertibility. For a general matrix $A\in \mathbb{C}^{m \times N}$, the pseudoinverse (also known as the Moore-Penrose inverse) is the uniquely determined matrix $A^\dagger \in \mathbb{C}^{N \times m}$ that satisfies the following four properties, the so-called Moore-Penrose conditions:

\begin{itemize}
    \item[1.] $A A^\dagger A  = A$
    \item[2.] $A^\dagger A A^\dagger= A^\dagger$
    \item[3.] $(A A^\dagger)^* = A A^\dagger$
    \item[4.] $(A^\dagger A)^* = A^\dagger A$
\end{itemize}

where $A^*$ denotes the adjoint matrix of $A$. The pseudoinverse  computed using various techniques, such as singular value decomposition. The pseudoinverse is employed in numerous applications, particularly in solving linear systems of equations that may not have a unique solution or might not have any solution. It is also utilized in data analysis and machine learning, particularly in linear regression and feature selection contexts.

Moreover, the pseudoinverse has applications in quantum computing, such as quantum error correction, quantum state preparation, and quantum machine learning. In general, the pseudoinverse is a valuable tool in quantum algorithms and holds the potential to enable significant speedups across a wide range of applications. Specifically, the pseudoinverse is essential for quantum algorithms since it can be used to solve linear systems of equations, which frequently arise in quantum computing. The HHL algorithm, developed by Harrow, Hassidim, and Lloyd in their groundbreaking work, is a quantum algorithm for solving linear systems of equations by computing the pseudoinverse.

The HHL algorithm is crucial because it offers a way to accelerate certain classical algorithms, such as solving linear systems of equations using Gaussian elimination, which is employed in various applications, including machine learning, optimization, and simulation. By utilizing the HHL algorithm, quantum computers can potentially solve these types of problems faster than classical computers. The HHL algorithm's runtime depends on the condition number, meaning the algorithm's efficiency increases with decreasing condition numbers. For small condition numbers, it has been demonstrated that the algorithm is exponentially faster than classical algorithms. However, determining the HHL algorithm's efficiency for a specific inverse problem requires prior knowledge of the condition number, which is calculated on digital hardware.

In \cite{Boche2022NonComputabilityOT}, the authors demonstrated, among other things, that the operator mapping a given matrix to its pseudoinverse or condition number is not Banach-Mazur computable and therefore, in general, not computable on digital hardware. Consequently, calculating the condition number of a matrix for the HHL algorithm does not provide reliable information about the efficiency of the HHL algorithm for the specific inverse problem.

\subsection{Differential Equations}\label{se:PDEs}


Differential equations are mathematical expressions that describe the relationship between a function and its derivatives concerning one or more independent variables. In other words, a differential equation is an equation involving the derivatives of a function. Differential equations naturally arise in various science and engineering fields and are used to model numerous physical phenomena, such as motion, heat transfer, fluid flow, and chemical reactions as well as phenomena in quantum mechanics. The Laplace equation, heat equation, and wave equation are the best-known and most studied equations. Differential equations are also employed in other fields like economics, finance, and biology. They can be classified into different categories based on properties such as order, linearity, and coefficient types. Finding solutions to various differential equation types is an essential topic in mathematics, with methods including analytical techniques like the separation of variables and the method of integrating factors, as well as numerical methods like Euler's method and the Runge-Kutta method. The runtime of numerical methods varies from one method to another and from one equation to another, usually expressed in terms of discretization parameters.

Despite its importance, the question of the computability and computational complexity of solutions on digital computers has largely been ignored in the literature. Thus, it was shown in  \cite{Boche2020TuringMC} that the wave equation 
\begin{equation}
    u_{tt}-c\Delta u=0
\end{equation} has, under certain conditions, classical solutions that are nowhere Turing computable. 
Furthermore, it has been shown in \cite{Bacho2022ComplexityBF} that the solutions of the Laplace equation
\begin{equation}
    \Delta u=0
\end{equation} and the heat equation
\begin{equation}
    u_{t}-\alpha\Delta u=0
\end{equation} are as hard to compute as a $\#P$-complete problem.
A similar complexity result has been shown in \cite{Boche2021ComplexityBI} for systems of ordinary differential equations. 

We will see in Section \ref{se:QuantumAnalog} that both, the non-computability result as well as the complexity result pose problems that can most likely not be solved with a digital quantum computer. 

\section{Possibilities of Quantum and Analog Computing Models}
\label{se:QuantumAnalog} 
The study of analog and quantum computers is an exciting area of research, especially when considering the potential advancements they could bring to the field of computing technology. Historically, analog computing, which depends on continuous physical variables such as voltage and current, has played a crucial role in scientific research and engineering applications. However, with the emergence of digital computing and Moore's Law, digital computing has become the prevailing paradigm. In contrast, quantum computing, which leverages quantum mechanical phenomena like superposition and entanglement to perform calculations, has gained interest in recent years due to its ability to solve specific problems exponentially faster than classical computers. As a result, there has been a substantial increase in research efforts dedicated to developing practical quantum computing technologies.

This section aims to explore the potential applications and possibilities of both analog and quantum computers, comparing their strengths and weaknesses to digital computers. It will investigate their distinct computational architectures and the types of problems they are most capable of solving. Additionally, the current state of research in developing these technologies will be examined. Through this exploration, we hope to gain a deeper understanding of how analog and quantum computing will contribute to shaping the future of computing technology. We refer the interested reader to \cite{Acn2018TheQT} and the references therein for a comprehensive roadmap of quantum technology in Europe. 

\subsection{Quantum Computability} \label{se:quantum.computability} 



Quantum algorithms, particularly those related to artificial intelligence, are usually designed as quantum circuits, an approach introduced by Deutsch \cite{Deutsch1989QuantumCN}. Quantum circuits consist of quantum gates that execute specific quantum operations on qubits. In these circuits, qubits are initialized in a known state before gates are applied to them in a particular sequence, transforming the qubits into their final state. The final state of a qubit encodes the computation's outcome. Quantum circuits are often visualized as diagrams where each gate is symbolized by a distinct icon, and qubits are depicted as lines. The flow of information is indicated by the lines' direction, and the order in which gates are applied is illustrated from left to right. Quantum circuits serve as a fundamental tool in quantum computing and are employed to design and implement quantum algorithms for various applications such as cryptography, optimization, and simulation.

Another theoretical model of computation is provided by the quantum Turing machine, also introduced by Deutsch \cite{Deutsch85QTCT}. This model can be considered a generalization of the classical Turing machine, enhanced with quantum gates and qubits. The quantum Turing machine operates on an infinite tape of qubits, which can exist in a superposition of states. It can perform quantum operations on these qubits, enabling it to solve specific problems more rapidly than classical computers.

In 1993, Yao \cite{Yao1993QuantumCC} demonstrated that quantum circuits and quantum Turing machines are computationally equivalent, signifying that they can simulate one another in polynomial time. This finding suggests that the two computational models possess equal computational power, which implies that any problem that can be efficiently solved by one model can also be efficiently solved by the other. Consequently, either computational model can be employed to solve quantum algorithms and execute quantum machine learning tasks.

In the same year, Bernstein and Vazirani \cite{Bernstein1993QuantumCT} established the existence of an efficient universal quantum Turing machine based on Deutsch's quantum Turing machine. Analogous to the classical universal Turing machine simulating the classical Turing machine, the universal quantum Turing machine can simulate an arbitrary quantum Turing machine on any input. However, constructing the universal quantum Turing machine is considerably more challenging than the classical universal Turing machine. Additionally, their work lends strong support to the notion that the quantum Turing machine should be considered a discrete computational model rather than an analog model, which is further reinforced by their results.

Moreover, it has been shown that the classical Turing machine can simulate qubits and vice versa, as exemplified in \cite[p. 202]{NieChu11QCQI}. This finding indicates that the class of problems solvable by a quantum Turing machine with unlimited time and space resources is no larger than the class of problems solvable by a classical Turing machine. Hence, by equivalence, both computational models share the same set of non-computable problems. Thus, there holds \\\\
\emph{A function is not computable on a classical Turing machine if and only if it is not computable on a quantum Turing machine.}\\

As a result, the same non-computability findings discussed in the previous sections also apply to quantum computers based on the quantum Turing machine. Owing to this fundamental issue, digital quantum computers are unable to overcome the problems faced by digital computers, see Section \ref{se:limitations}.

\subsection{Quantum Complexity} \label{se:quantum.complexity} 

Regarding complexity, the quantum Turing machine has demonstrated its superiority over classical computers in certain cases. Quantum computers operate on the principles of quantum mechanics, allowing them to perform specific calculations much faster than classical computers. A crucial advantage of quantum computers is their ability to conduct multiple calculations simultaneously, using a property known as superposition. This means a quantum computer can evaluate many potential solutions to a problem concurrently, as opposed to classical computers which assess solutions one at a time.

Another essential property of quantum computers is entanglement, enabling multiple qubits (quantum bits) to connect so that their states are strongly correlated, even when they are physically distant. This correlation can be harnessed to execute specific computations more efficiently than classical computers. Quantum computers excel at solving particular types of problems, such as factorization and search problems. For instance, factoring large numbers is challenging for classical computers, but quantum computers can efficiently factor large numbers using Shor's algorithm \cite{Shor1995PolynomialTimeAF}.

Quantum computers hold potential value for a variety of applications, from cryptography to drug discovery. However, it remains uncertain whether quantum computers can provide a significant speedup advantage over classical computers for all problem types. It is believed that quantum computers have a quantum advantage for certain computational tasks, with the scope of this advantage depends on the specific problem and the quantum computer's characteristics.

For specific problems such as factorization and discrete logarithm, quantum computers are known to offer an exponential speedup over classical algorithms. This means that the time required to solve the problem decreases exponentially as the problem size increases. Factorization and discrete logarithm problems are relevant for cryptography and information security. The development of quantum computers with a sufficient number of qubits could potentially undermine many of today's encryption methods. 

In other cases, such as optimization problems and quantum simulation, quantum computers are anticipated to provide a polynomial speedup over classical algorithms, meaning that the time required to solve the problem decreases polynomially as the problem size increases. While this may not yield the same exponential speedup as factorization, it can still significantly impact various scientific and engineering fields. Another example where quantum advantage has been demonstrated experimentally and theoretically is the quantum random sampling problem like the Gaussian boson sampling problem \cite{Zhong2021PhaseProgrammableGB, Aaronson2010TheCC,Hangleiter2022ComputationalAO}. 

It is crucial to recognize that not all problems will benefit from a quantum advantage. Some problems, like sorting and searching, can already be solved efficiently by classical computers and may not gain from a quantum approach. For example, Grover's algorithm \cite{Grover1996AFQ} only offers a quadratic speedup over a classical algorithm. Moreover, it has been shown that Grover's algorithm is asymptotically optimal, suggesting that quantum algorithms cannot efficiently solve NP-complete problems.

Complexity classes for quantum Turing machines can be introduced similarly to classical Turing machines, and understanding the relationship between these classes and classical complexity classes is an active area of research. For instance, Bernstein and Vazirani \cite{Bernstein1993QuantumCT} defined a complexity class called BQP (bounded-error quantum polynomial time), which includes decision problems that can be solved by a quantum computer using a polynomial number of gates and qubits, with a bounded probability of error. They demonstrated that some problems in BQP are difficult for classical computers. One such problem is Simon's problem, which involves finding a hidden string of bits given a black-box function that computes a one-to-one or two-to-one function of the input. The authors showed that a quantum algorithm can solve Simon's problem using a polynomial number of gates and qubits, while the best-known classical algorithm requires an exponential number of queries to the black-box function. This finding indicates that quantum computers can provide an exponential speedup over classical computers for specific problems.

However, by demonstrating the existence of a problem that can be solved in polynomial time on a quantum Turing machine but requires superpolynomial time on a bounded-error probabilistic Turing machine, the authors show that quantum Turing machines violate the modern formulation of the Church-Turing thesis. As mentioned before, this thesis states that a Turing-computable function can always be efficiently computed by a random-polynomial time algorithm.

\subsection{Quantum Machine Learning} \label{se:QML} 
Next, we consider the implications of the observations made in the previous sections for quantum machine learning (QML). Quantum Machine Learning is an interdisciplinary field of research that combines principles from quantum mechanics and machine learning. Its goal is to develop innovative algorithms and techniques to enhance the performance of machine-learning tasks by harnessing the unique properties of quantum systems. In traditional machine learning, data is processed and analyzed using classical algorithms that operate on classical bits, which can have values of 0 or 1. Conversely, quantum machine learning algorithms employ quantum bits, or qubits, which can exist in a superposition of 0 and 1 states, as well as entanglement, which enables correlations between qubits that are not achievable in classical systems.

QML algorithms can be broadly categorized into three groups: quantum-inspired, quantum-assisted, and fully quantum. Quantum-inspired algorithms are classical algorithms influenced by quantum mechanics, seeking to emulate some of their properties to enhance performance. Quantum-assisted algorithms combine classical machine learning methods with small quantum computations to expedite specific tasks. Lastly, fully quantum algorithms depend on large-scale quantum computations to execute machine learning tasks. Some examples of QML applications include image and speech recognition, recommendation systems, and optimization problems. However, QML is still an emerging field, with many of its applications under development and exploration. Nevertheless, the potential advantages of QML are substantial, potentially achieving exponential speedups over classical algorithms in some cases and overcoming certain limitations of classical machine learning methods. Despite the significant computational speedup, the general non-computability problem for neural networks persists, as indicated by Theorem \ref{th:QuantumComp} and the results presented in Section \ref{se:limitations}.

In the following section, we will discuss how analog computers, at least theoretically, offer a solution to this problem by performing computations with real numbers rather than solely with rational numbers.

\subsection{An Analog Computing Model - The Blum-Shub-Smale Machine} \label{se:BBS machine} 
Digital quantum computers do not resolve the issue of non-computability, as it is intrinsic to the discrete nature of the Turing machine. However, certain analog computing models show considerable potential in this regard. A well-known analog computing model is the Blum-Shub-Smale (BSS) machine, proposed by Blum, Shub, and Smale \cite{Blum1989OnAT} in the 1980s. The BSS machine aims to be a more realistic computational model than the Turing machine, considering the limitations of actual computing hardware by executing arithmetic operations on real numbers. The machine works with a set of real numbers and can perform basic arithmetic operations, such as addition, subtraction, multiplication, and division, as well as comparisons between real numbers. It also includes operations like rounding and truncation, essential for performing arithmetic on real numbers in actual hardware. The BSS machine is also employed as a theoretical tool to study the computational complexity of numerical algorithms and problems in cryptography and number theory. It has been applied to various issues in computational algebra, geometry, and optimization, among other areas. A crucial result in the study of the BSS machine is the BSS model of computation, which characterizes the complexity of numerical problems that the BSS machine can solve efficiently.
We have seen in Section \ref{se:inverse.problems} that the pseudo inverse is not Banach-Mazur computable on a Turing machine. However, an interesting finding presented in \cite{Boche2022InversePA} shows that the pseudo-inverse is computable on the Blum-Shub-Smale machine. This discovery suggests that operations that are impossible to perform on a digital computer can be accomplished on an analog computer that operates on the Blum-Shub-Smale computing model. Thus, BSS machines have been found to have strictly greater capabilities than Turing machines for solving inverse problems, when allowed to process real numbers exactly. However, it is important to note that these findings are based on a specific problem formulation, and it remains an open question whether these conclusions can be applied to a more general class of inverse problems.

The implications of these results are noteworthy for both analog and digital computing devices. In general, analog hardware modeled by BSS machines allows for a solver which can be applied to any inverse problem of a given dimension. In contrast, on digital hardware, the algorithmic solvability is severely limited, and thus approaches to solve real inverse problems on upcoming analog hardware in the form of neuromorphic devices may potentially have a strictly greater capacity than current digital solutions.

Even though exact computations are theoretically possible on an analog computer, as we will discuss in the next section, the practical realization of such computations is still a significant challenge due to physical limitations.

\section{Conclusion and Outlook}\label{se:outlook}

In this review, we have observed that digital hardware has notable limitations when it comes to executing exact computations on continuous quantities and resolving certain problem classes. This is because digital hardware cannot feasibly represent these quantities exactly. Consequently, merely approximate solutions to problems with continuous quantities can be achieved on digital hardware, and the quality of the approximation can only be assessed if the distance to the true (exact) solution is known. However, in many instances, the exact solution is not computable on a digital computer. This non-computability issue can lead to negative consequences in vital areas such as artificial intelligence, information theory, finance, cryptography, optimal transport, and linear programming, which play a crucial role in nearly all aspects of modern life.

Moreover, we have shown that this result applies not only to classical computers but also to quantum computers, which are based on the quantum Turing machine computing model. Despite this, we have also observed that the digital quantum Turing machine outperforms the classical Turing machine in terms of algorithmic complexity for specific problem classes. Quantum computers operate by manipulating quantum states, which can be in superposition or entanglement, enabling them to perform certain computations much faster than classical computers. Certain algorithms, like Shor's algorithm, exemplify this significant advantage. Conversely, for problem classes such as searching, the quantum computer offers only a quadratic speedup using the asymptotically optimal Grover algorithm. While quantum computers generally perform mathematical operations much faster than classical computers, they cannot resolve the non-computability problem if they are solely based on a discrete computing model.

One potential method to bypass these limitations is by employing analog computers or analog quantum computers. In Section \ref{se:BBS machine}, we have seen that, at least theoretically, the analog computing model of the BSS machine can address the non-computability problem for specific problem classes. Analog computing devices, like BSS machines, can store and process real numbers precisely without error, which is not possible on digital hardware. As a result, they can handle continuous quantities and perform exact computations. For instance, BSS machines can solve any inverse problem of a given dimension, while on digital hardware, algorithmic solvability is heavily restricted. However, these positive results are entirely theoretical.

Although there are ongoing efforts to develop practical analog computing devices, numerous physical limitations must be overcome before these devices can be fully utilized. To answer the initial question of whether the next generation of artificial intelligence requires quantum computing: yes, the next generation necessitates quantum computing in both digital and analog forms to overcome theoretical limitations. The appearance and usage of these quantum and analog computers in the future are still open questions and are subjects of active research on both practical and theoretical levels.

\section*{Acknowledgments}

This work of G. Kutyniok and H. Boche was supported by LMUexcellent and TUM AGENDA 2030, funded by the Federal Ministry of Education and Research (BMBF) and the Free State of Bavaria under the Excellence Strategy of the Federal Government and the Länder as well as by the Hightech Agenda Bavaria.

G. Kutyniok acknowledges support from the Konrad Zuse School of Excellence in Reliable AI (DAAD), the Munich Center for Machine Learning (BMBF) as well as the German Research Foundation under Grants DFG-SPP-2298, KU 1446/31-1 and KU 1446/32-1 and under Grant DFG-SFB/TR 109, Project C09 and the Federal Ministry of Education and Research under Grant MaGriDo.

This work of H. Boche was supported in part by the German Federal Ministry of Education and Research (BMBF) under Grant 16ME0442. H. Boche was also supported in part by the German Federal Ministry of Education and Research within the national initiative on 6G Communication Systems through the research hub 6G-life under Grants 16KISK002.





\bibliographystyle{my_alpha}
\bibliography{bib_aras}

%
%

%

\end{document}